\documentclass[lettersize,journal]{IEEEtran}
\usepackage{amsmath,amsfonts}
\usepackage[linesnumbered,ruled,vlined]{algorithm2e}

\SetNlSty{textbf}{\color{gray}}{}   
\SetAlgoNlRelativeSize{-1}
\SetNlSkip{0.6em}
\bstctlcite{BSTcontrol:NoDash}

\usepackage{array}
\usepackage[caption=false,font=normalsize,labelfont=sf,textfont=sf]{subfig}
\usepackage{textcomp}
\usepackage{stfloats}
\usepackage{stfloats}
\usepackage{url}
\usepackage{bm}
\usepackage{mathrsfs} 
\usepackage{verbatim}
\usepackage{graphicx}
\usepackage{url}
\usepackage{verbatim}
\usepackage{graphicx}
\usepackage{cleveref}
\usepackage{array, multirow, booktabs}

\usepackage[table]{xcolor}
\definecolor{rowcolor}{rgb}{0.898,0.949,0.969}
\usepackage{stfloats}
\usepackage{url}
\usepackage{verbatim}
\usepackage{mathrsfs}
\usepackage{graphicx}
\usepackage{multirow}
\usepackage{makecell}
\usepackage{cleveref}
\usepackage{bm}
\usepackage{booktabs}
\usepackage{pifont}
\usepackage{tikz}
\usepackage{bigstrut}

\usepackage{amssymb}
\usepackage[table]{xcolor}
\definecolor{rowcolor}{rgb}{0.898, 0.949, 0.969}
\Crefname{figure}{Fig.}{Figs.}
\usepackage[letterpaper, pass]{geometry}
\usepackage{titlesec}

\usepackage{cite}
\begin{document}

\title{\vspace{18pt}Spatial-VLN: Zero-Shot Vision-and-Language Navigation With Explicit Spatial Perception and Exploration}

\author{Lu Yue$^{\dagger}$, Yue Fan$^{\dagger}$, Shiwei Lian, Yu Zhao, Jiaxin Yu, Liang Xie$^{*}$ and Feitian Zhang$^{*}$

\thanks{\textsuperscript{\textdagger} These authors contributed equally to this work.}%
\thanks{L. Yue, Y. Fan, S. Lian and F. Zhang are with the Robotics and Control Laboratory, School of Advanced Manufacturing and Robotics, and the State Key Laboratory of Turbulence and Complex Systems,  Peking University, Beijing, 100871, China. L. Yue is also with the Defense Innovation Institute, Academy of Military Sciences, Beijing 100071, China, and Tianjin Artificial Intelligence Innovation Center, Tianjin 300450, China.}%
\thanks{Y. Zhao is with the Institute of Computing and Intelligence, Harbin Institute of Technology (Shenzhen), Shenzhen, 518000, China.}
\thanks{J. Yu and L. Xie are with the Defense Innovation Institute, Academy of Military Sciences, Beijing, 100071, China, and Tianjin Artificial Intelligence Innovation Center, Tianjin, 300450, China.}
\thanks{Corresponding authors are L. Xie and F. Zhang (email: feitian@pku.edu.cn).} %
}





\maketitle

\begin{abstract}
Zero-shot Vision-and-Language Navigation (VLN) agents leveraging Large Language Models (LLMs) excel in generalization but suffer from insufficient spatial perception. Focusing on complex continuous environments, we categorize key perceptual bottlenecks into three spatial challenges: door interaction, multi-room navigation, and ambiguous instruction execution, where existing methods consistently suffer high failure rates.
We present Spatial-VLN, a perception-guided exploration framework designed to overcome these challenges. The framework consists of two main modules. The Spatial Perception Enhancement (SPE) module integrates panoramic filtering with specialized door and region experts to produce spatially coherent, cross-view consistent perceptual representations. 
Building on this foundation, our Explored Multi-expert Reasoning (EMR) module uses parallel LLM experts to address waypoint-level semantics and region-level spatial transitions.
When discrepancies arise between expert predictions, a query-and-explore mechanism is activated, prompting the agent to actively probe critical areas and resolve perceptual ambiguities.
Experiments on VLN-CE demonstrate that Spatial-VLN achieves state-of-the-art performance using only low-cost LLMs. Furthermore, to validate real-world applicability, we introduce a value-based waypoint sampling strategy that effectively bridges the Sim2Real gap. Extensive real-world evaluations confirm that our framework delivers superior generalization and robustness in complex environments. Our codes and videos are available at https://yueluhhxx.github.io/Spatial-VLN-web/.
\end{abstract}


\section{Introduction}
In Vision-and-Language Navigation in Continuous Environments (VLN-CE) tasks, agents interpret natural language instructions to execute low-level actions toward a target without prior access to topological maps. Unlike discrete settings, VLN-CE demands robust environmental perception and decision-making. This is necessary to overcome visual discrepancies caused by simulation reconstruction errors or sim-to-real gaps, as well as the redundancy inherent in continuous 
\begin{figure}[t]
  \centering
   \includegraphics[width=1\linewidth]{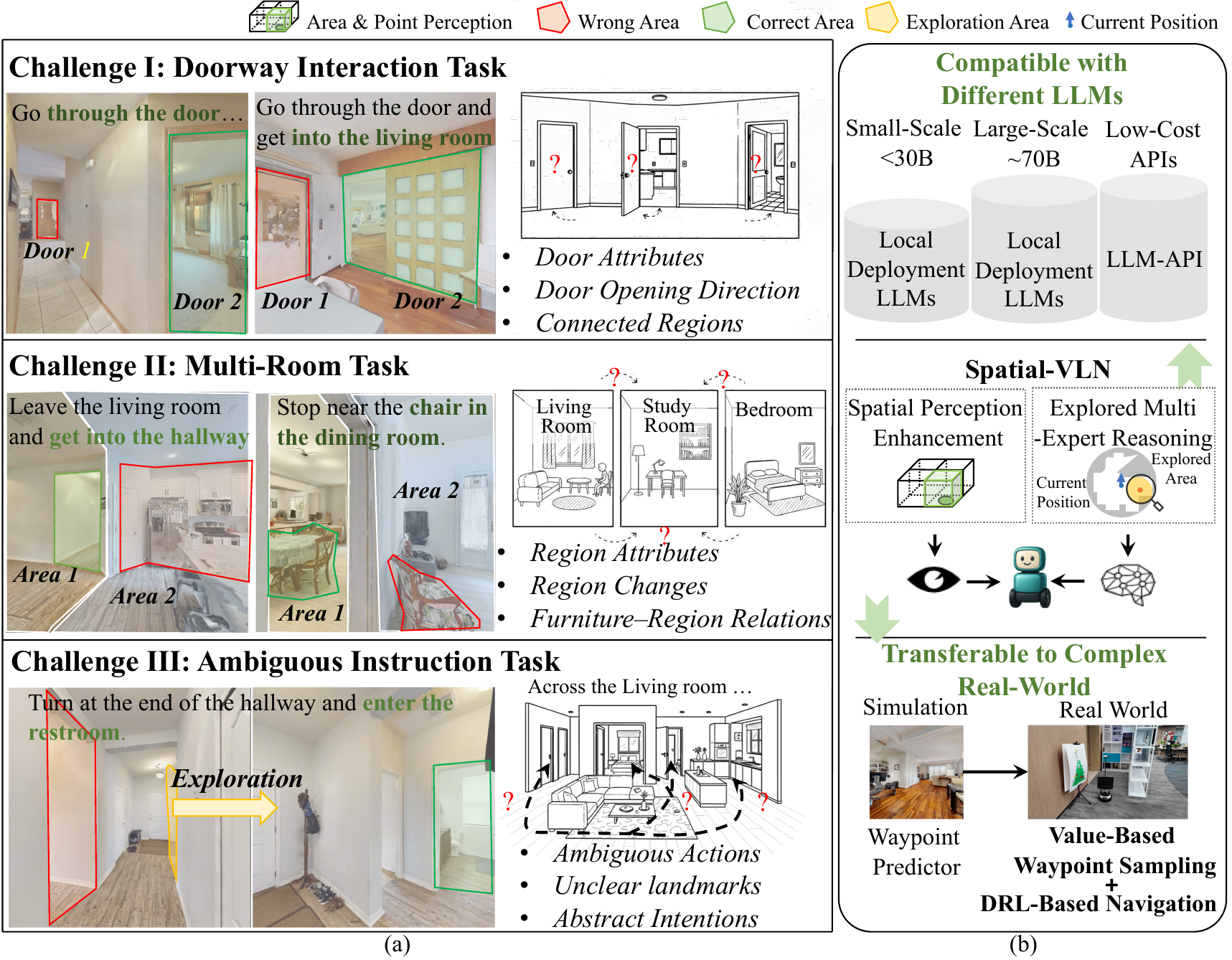}
   \caption{{Overview of challenging spatial tasks and our proposed Spatial-VLN framework.
(a) Three spatial challenging tasks: (1) Doorway Interaction – the agent must reason about door state (open/closed), orientation, and semantic transitions between connected regions; (2) Multi-room Transition – the instruction involves navigating through multiple large or ambiguous regions, requiring recognition of region boundaries and semantic shifts; (3) Landmark-Sparse Navigation – the instruction lacks explicit landmarks, forcing the agent to infer spatial intent based on vague directional cues and minimal semantic guidance. (b) The proposed framework Spatial-VLN contains spatial perception enhancement module and explored multi-expert reasoning module. Our approach is validated using various LLM experts and demonstrates robust transferability to real-world environments.}}
   \label{fig:motivation}
\end{figure}actions~\cite{krantz2020beyond,anderson2021sim}.
Many existing approaches rely on end-to-end models, which learn latent visual-linguistic alignments through extensive training~\cite{chen2022weakly,an2024etpnav,an2023bevbert}. However, they typically suffer from high computational costs and poor generalization to unseen real-world scenes. Alternatively, instead of task-specific fine-tuning, zero-shot approaches utilize Large Language Models (LLMs) and Vision-Language Models (VLMs) for high-level task planning ~\cite{ren2024embodied, liu2025aligning} while employing learning-based methods for robust low-level motion control~\cite{zhu2021rule, zhang2022ipaprec,lian2024transferability}. It enables agents to achieve generalizable navigation by directly leveraging the vast world knowledge embedded in LLMs~\cite{chen2025constraint, shi2025smartway, long2024instructnav, qiao2025open}.
Despite their semantic reasoning prowess, these methods face a critical bottleneck regarding fine-grained spatial perception. LLMs inherently struggle to ground abstract linguistic instructions into physical spatial constraints. Consequently, they frequently overlook environmental cues such as geometric layouts, regional connectivity, and object orientations.

To systematically quantify this perceptual gap, we comprehensively analyze typical failure cases in existing methods and categorize them into three representative spatial challenge tasks: doorway interaction, multi-region and ambiguous instruction tasks. These tasks represent distinct levels of spatial complexity spanning from geometric interaction and spatial change to abstract environmental reasoning. For instance, as illustrated in \Cref{fig:motivation}, simple commands such as `go through the door and get into the living room' frequently fail in scenarios with multiple visible doors. This failure stems not from semantic misunderstanding but from a lack of spatial awareness regarding door orientation or the spatial relationship between regions. Neglecting these spatial constraints leads to significant performance degradation in complex scenarios.

To address the aforementioned challenges, we propose Spatial-VLN, a perception-guided exploration framework designed to inject explicit spatial awareness into zero-shot navigation. Our framework comprises two core modules. First, the Spatial Perception Enhancement (SPE) module mitigates single-view inconsistency by fusing multi-sensor data to explicitly identify door attributes and region boundaries. Second, we introduce an Explored Multi-expert Reasoning (EMR) module which orchestrates a waypoint expert for object-level alignment and a region expert for region transitions. Crucially, a conflict-driven exploration mechanism is triggered whenever expert predictions diverge. This mechanism guides the agent toward informative areas to resolve decision ambiguity through supplementary observations.
{Finally, to validate our approach in practice, we developed a sim-to-real deployment strategy tailored for real-world transitions, which focuses on candidate waypoint selection and low-level navigation execution. To address common real-world failures where predictors generate unnavigable points or suffer from insufficient semantic information, we propose a value-based sampling method for robust waypoint selection. Meanwhile, by integrating a multi-sensor fusion map with a reinforcement learning-based controller, our framework achieves real-time obstacle avoidance.}

In summary, our main contributions are threefold. First, we identify and formalize three critical spatial challenges in VLN-CE, highlighting the necessity of explicit spatial reasoning. Second, we propose the Spatial-VLN framework, which integrates panoramic spatial perception with a conflict-driven multi-expert reasoning mechanism, achieving state-of-the-art (SoTA) performance with low-cost LLM experts. Finally, we demonstrate the robust sim-to-real transferability of our method through extensive real-world experiments, validating the effectiveness of Spatial-VLN in complex environments.

\section{Related Work}
\subsection{Traditional VLN-CE}
VLN-CE requires an agent to execute low-level actions to reach a target described by natural language, without prior access to topological maps. Unlike discrete environments, VLN-CE operates within reconstructed simulations or the physical real world, typically utilizing waypoint predictors~\cite{hong2022bridging} to estimate candidate distributions based on visual inputs. This setting imposes stringent perceptual demands due to significant environmental visual noise, the absence of fixed topological structures, and often restricted or suboptimal viewpoints, which frequently result in ambiguous semantic information.

Existing approaches primarily adopted end-to-end learning paradigms, training models on extensive datasets to learn direct mappings from observations to actions~\cite{wang2023scale,wang2023gridmm,hong2024navigating}. However, these methods suffer from high computational costs and poor generalization. To address these limitations, recent research has pivoted towards leveraging the vast world knowledge and strong semantic reasoning capabilities of LLMs as high-level navigation brains. Some works adopt hierarchical architectures, integrating LLMs with waypoint-based planning where the model generates subgoals, while perception modules handle local grounding~\cite{hong2022bridging}. Other studies focus on augmenting LLMs with semantic knowledge to better capture object-level cues for decision-making~\cite{wu2024vision, an2021neighbor}.

\subsection {Zero-shot VLN-CE }
Zero-shot VLN enables agents to follow linguistic instructions in unseen environments without task-specific training, leveraging the generalization capabilities of foundation models. The core challenge lies in effectively adapting the general knowledge of LLMs to the complex embodied navigation tasks. 
Recent approaches typically structure zero-shot VLN-CE into modular reasoning and planning pipelines. For instance, {some frameworks formulate navigation as sequential sub-instruction execution with constraint-aware progress tracking~\cite{chen2025constraint}}, while others focus on obstacle-aware waypoint prediction and history-based trajectory refinement~\cite{li2025boosting, shi2025smartway}. To enhance adaptability, researchers have also explored constructing simplified scene representations for efficient planning~\cite{bhatt2025vln} or employing unified value-maps to support diverse instruction types~\cite{long2024instructnav}. Recognizing the bottlenecks of single-model systems, DiscussNav~\cite{long2024discuss} introduces multi-expert discussion strategies to reduce reasoning errors, while Open-Nav~\cite{qiao2025open} demonstrates that open-source LLMs achieve competitive performance when equipped with structured prompts.

However, most existing frameworks predominantly rely on two-dimensional semantic recognition, lacking an explicit understanding of the environment's 3D spatial structure. This deficiency in spatial perception renders these methods vulnerable to hallucinations and cumulative errors when facing complex spatial environments. To address this, we investigate the spatial challenges in VLN-CE and propose Spatial-VLN, a spatially-aware multi-expert architecture. By integrating multi-sensor fusion to acquire explicit spatial perception and introducing a perception guided exploration mechanism, our approach enables the agent to actively resolve decision conflicts through environmental interaction, ensuring robust navigation.

\section{Method}

\subsection{Problem Formulation}
\label{sec:problem}
In VLN-CE setting, each navigation episode provides the agent with a natural-language instruction $\mathbf{I}$.
During navigation, the agent continuously maintains its positional state and acquires panoramic RGB-D observations at every time step \(t\). Each observation is formulated as
$\mathbf{O}_t = \{(\mathbf{o}^{\text{rgb}}_{t,i}, \mathbf{o}^{\text{d}}_{t,i})\}_{i=1}^{12}$,
containing 12 viewpoints uniformly distributed along the horizontal plane, with viewing angles defined by \(\theta_i = 30^\circ \times (i-1)\). Here, $\mathbf{o}^{\text{rgb}}_{t,i}$ and \(\mathbf{o}^{\text{d}}_{t,i} \) represent the RGB and depth images captured from the \(i\)-th direction, respectively.
To reach the target specified by the instruction, the agent must effectively integrate textual information with multi-view visual observations. Unlike waypoint-based navigation, VLN-CE requires the agent to operate directly within a continuous 3D environment. Accordingly, the agent chooses actions from the following low-level control set:
$\mathscr{A} = \{\text{forward 0.25\,m},\ \text{turn left }15^\circ,\ \text{turn right }15^\circ,\ \text{stop}\}$.
To acquire environmental perception, we employ RAM~\cite{zhang2024recognize} to extract semantic features and utilize Spatialbot~\cite{cai2025spatialbot} to generate spatial semantic descriptions. Furthermore, we leverage Large Language Models as `Experts' to integrate these environmental observations with task instructions to reason the next action.

\begin{figure*}[!t]
\centering
\includegraphics[width=1\linewidth]{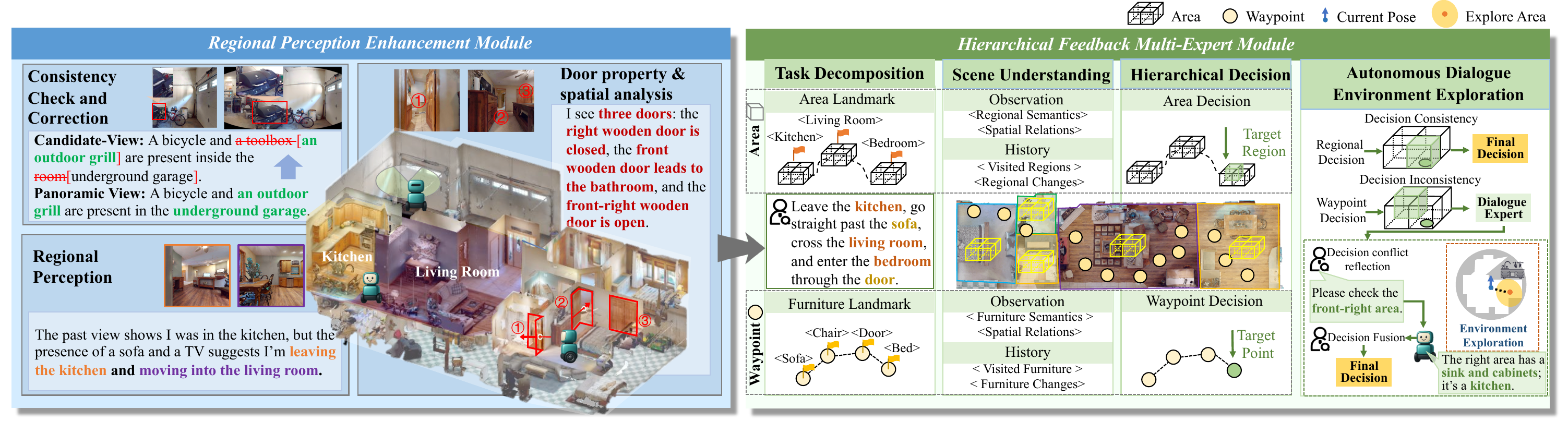}
\caption{The architecture of the proposed Spatial-VLN framework. The framework comprises two core components: the Spatial Perception Enhancement (SPE) module, which generates spatially consistent semantic descriptions while enriching spatial attributes regarding doors and regional transitions, and the Explored Multi-expert Reasoning (EMR) module, which executes dual-expert reasoning based on the enhanced perception. By aligning waypoint-level and region-level inferences, the EMR module triggers an active exploration mechanism to resolve ambiguities whenever decision inconsistencies arise.}
\label{fig:main_method}
\end{figure*}

\subsection{Spatial-Challenging Tasks}


Although LLM-based zero-shot approaches demonstrate proficiency in linguistic interpretation, their embodied navigation suffers from spatial perception deficiencies that hinder the accurate extraction of environmental cues. {To systematically analyze this limitation}, we summarize common failures into three representative spatial challenges. 
Empirical evaluations in \Cref{sec:exp} reveal that SoTA algorithms exhibit significant performance degradation across these scenarios, identifying them as critical areas for enhancing navigation robustness.

\textit{Doorway Interaction Challenges.} 
Unlike static objects, doors possess dynamic properties, including opening states and regional connectivity, which directly dictate navigability. Failure to perceive these cues often results in critical errors, such as bypassing necessary transitions or entering incorrect rooms. Our analysis reveals that in tasks involving frequent door-related descriptions, the success rate of the baseline drops by 6\%, highlighting the significant challenge these scenarios pose to existing methods.

\textit{Multi-Room Challenges.} 
Navigating across multiple rooms requires the continuous tracking of spatial boundaries. We observe a 2\% drop in baseline success rate as the number of referenced regions increases. This decline primarily stems from existing strategies prioritizing furniture-level semantics over region-level cues. Consequently, this insufficient region awareness leads to reasoning errors, including searching within unintended rooms or misinterpreting area sequences.

\textit{Ambiguous Instruction Challenges.} 
Real-world instructions frequently lack explicit visual anchors. Empirical results demonstrate a 4\% decline in baseline performance as the density of concrete landmarks decreases. This decline is attributed to the increased difficulty of inferring implicit spatial relationships without clear directional cues. This reasoning process is prone to errors, often failing to ground abstract descriptions, which results in landmark misidentification and navigation failure.

\begin{figure}[!t]
\centering
\includegraphics[width=1\linewidth]{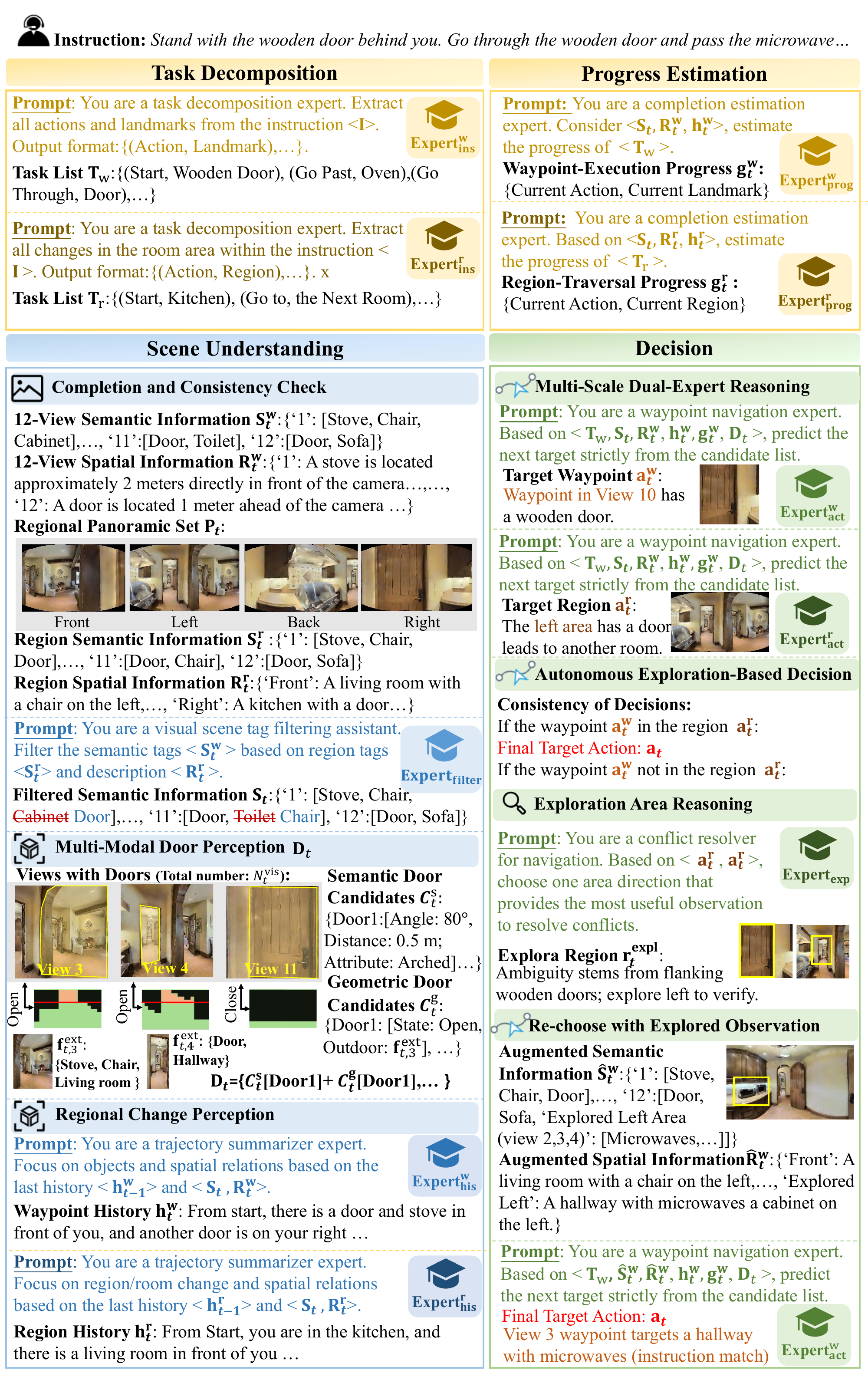}
\caption{Illustration of prompts for various LLM-based experts within Spatial-VLN. The prompts empower specialized experts in task decomposition, progress estimation, perception filtering, and history summarization, as well as those focused on exploration and reasoning, to perform critical sub-tasks from key perception extraction to final action decision-making.}
\label{expert}
\end{figure}

\subsection{Spatial-VLN}\label{sec:main}

As illustrated in \Cref{fig:main_method}, we propose Spatial-VLN, a zero-shot framework mitigating spatial perception deficits in LLM agents via two core modules. The SPE module extracts consistent spatial semantic information of regions and doors, while the EMR module employs a dual-expert architecture to align furniture and region-level decisions. To resolve environmental uncertainty, an active exploration mechanism is triggered by decision conflicts. \Cref{expert} further details the processing workflow, expert input-output specifications, and prompt designs through a representative case.

\noindent\textbf{Spatial Perception Enhancement Module.}
To augment the decision-making stage with granular environmental context, this module prioritizes the extraction of navigation-critical spatial features. Specifically, we leverage semantic recognition and descriptive analysis on stitched panoramic images to generate spatially consistent representations, and focus on the structural attributes of doors and functional regions.

\textit{Consistency Check and Correction.}
Accurate semantic recognition is foundational to effective spatial perception. 
At each timestep $t$, given the observation $\mathbf{O}_t=\{(\mathbf{o}^{\text{rgb}}_{t,i},\,\mathbf{o}^{\text{d}}_{t,i})\}_{i=1}^{12}$, the RAM~\cite{zhang2024recognize} and Spatialbot~\cite{cai2025spatialbot} generate initial single-view semantic predictions $\mathbf{S}^{\text{w}}_t = \text{RAM}(\mathbf{O}_t) = \{\mathbf{s}^{\text{w}}_{t,i}\}_{i=1}^{12}$ and spatial relations $\mathbf{R}^{\text{w}}_t = \text{Spatialbot}(\mathbf{O}_t) =\{\mathbf{r}^{\text{w}}_{t,i}\}_{i=1}^{12}$, respectively, where superscripts `w' and `r' denote waypoint-level and region-level operations.
However, predictions derived from isolated views often suffer from inconsistency due to limited fields of view and occlusions. 
To mitigate this, we partition the twelve views into four cardinal regions $\mathcal{R}=\{\text{Front}, \text{Left}, \text{Back}, \text{Right}\}$, covering consecutive $90^\circ$ angular intervals. 
For each region, we synthesize a cohesive panorama $\mathbf{p}_{t,j}$ via feature-based image stitching~\cite{brown2007automatic}, yielding a regional panoramic set $\mathbf{P}_t=\{\mathbf{p}_{t,j}\}_{j=1}^4$. 
Subsequently, RAM is applied to $\mathbf{P}_t$ to obtain a consistent region semantic reference $\mathbf{S}^{\text{r}}_t = \{\mathbf{s}^{\text{r}}_{t,j}\}_{j=1}^{4}$. 
Leveraging this panoramic context, a filtering expert refines the initial predictions by discarding semantics in $\mathbf{S}^{\text{w}}_t$ that diverge from $\mathbf{S}^{\text{r}}_t$, yielding the final semantic output $\mathbf{S}_t = \text{Expert}_{\text{filter}}(\mathbf{S}^{\text{w}}_t, \mathbf{S}^{\text{r}}_t)$. 
Finally, Spatialbot processes $\mathbf{P}_t$ to derive the comprehensive spatial representation $\mathbf{R}^{\text{r}}_t=\{\mathbf{r}^{\text{r}}_{t,j}\}_{j=1}^{4}$. 
This panoramic approach transcends the limitations of single-view analysis, guaranteeing spatial reasoning completeness and coherence.

\textit{Regional Perception.}
Since instructions often prescribe sequences of areas, accurately detecting region transitions is vital for tracking task progress. 
To capture the temporal dynamics of these transitions, we introduce a dual-scale history mechanism comprising two specialized experts. 
First, the waypoint history expert, denoted as $\text{Expert}^{\text{w}}_{\text{his}}$, focuses on fine-grained local semantics and orientation changes. 
It updates the waypoint history state $\mathbf{h}^{\text{w}}_t$ by integrating the previous state $\mathbf{h}^{\text{w}}_{t-1}$ with the current waypoint-level perception
    $\mathbf{h}^{\text{w}}_t = \text{Expert}^{\text{w}}_{\text{his}}(\mathbf{h}^{\text{w}}_{t-1},\, \mathbf{S}_t, \mathbf{R}_t^{\text{w}})$.
Complementarily, the region history expert, $\text{Expert}^{\text{r}}_{\text{his}}$, targets broader regional contexts and global orientation shifts. 
It maintains the region history state $\mathbf{h}^{\text{r}}_t$ based on panoramic spatial inputs 
    $\mathbf{h}^{\text{r}}_t = \text{Expert}^{\text{r}}_{\text{his}}(\mathbf{h}^{\text{r}}_{t-1},\, \mathbf{S}^{\text{r}}_t, \mathbf{R}^{\text{r}}_t)$.
These multi-scale history experts individually capture dynamics at the waypoint and region levels, enabling the agent to reason with multi-scale fine-grained information.

\textit{Multi-Modal Door Perception.}
To ensure precise interaction, we propose a fusion approach combining semantic understanding (superscript $\text{s}$) with geometric verification (superscript $\text{g}$). We first identify $N^\text{s}_t$ door views from predictions $\mathbf{S}_t$, where for each view $i$, a region of interest (ROI) $\mathbf{b}_{t,i}$ is defined. By employing Spatialbot with door-specific prompts, we estimate the distance $d_{t,i}$ and appearance attributes $\mathbf{a}_{t,i}$ (e.g., material, color) to form the visual candidate set $\mathcal{C}_t^\text{s} = \{(\theta_{t,i}^\text{s}, \mathbf{b}_{t,i}, d_{t,i}, \mathbf{a}_{t,i})\}_{i=1}^{N^\text{s}_t}$, where $\theta_{t,i}^\text{s}$ represents the relative azimuth derived from the ROI center. Subsequently, geometric openings $\mathcal{C}_t^\text{g} = \{\theta_{t,j}^\text{g}\}_{j=1}^{N^\text{g}_t}$ are extracted from the LiDAR occupancy map by detecting sharp depth discontinuities (see \Cref{expert}). A visual candidate $i$ is classified as `open' ($\mathbf{u}_{t,i}=\text{`open'}$) only if it spatially aligns with a geometric opening within a specified angular tolerance; otherwise, it is labeled `closed.' This cross-verification filters out architectural features like corridors that mimic door geometry but lack semantic identity. For confirmed `open' doors, a secondary semantic recognition is performed within $\mathbf{b}_{t,i}$ to identify the external room type or furniture $\mathbf{f}_{t,i}^\text{ext}$ beyond the confidence threshold. The final comprehensive door state is formalized as $\mathbf{D}_t = \{(\theta_{t,k}^\text{s}, d_{t,k}, \mathbf{a}_{t,k}, \mathbf{u}_{t,k}, \mathbf{f}_{t,k}^\text{ext})\}_{k=1}^{N_t}$, where $N_t$ is the number of successfully verified doorways at time $t$.

\noindent\textbf{Explored Multi-Expert Reasoning Module.}
Although the perception module ensures semantic consistency, it remains insufficient for complex instruction-to-action aligning. To bridge this gap, multi-scale reasoning with uncertainty-driven exploration is introduced. Specifically, waypoint and region experts collaborate to predict targets based on local semantics and region changes. Moreover, decision discrepancies between two experts trigger autonomous exploration to retrieve missing visual cues, resolving ambiguity and ensuring robust navigation under occlusion.

\textit{Multi-Scale Dual-Expert Reasoning.}
Navigation instructions inherently encompass both fine-grained object interactions and high-level regional transitions. 
To address this duality, we decompose instruction $\mathbf{I}$ into parallel waypoint and region sequences: $\mathbf{T}_{\text{w}}=\{(\text{Act}^{\text{w}}_{n_{\text{w}}}, \text{Fur}_{n_{\text{w}}})\}_{n_{\text{w}}=1}^{{N_{\text{w}}}}$ and $\mathbf{T}_{\text{r}}=\{(\text{Act}^{\text{r}}_{n_{\text{r}}}, \text{Area}_{n_{\text{r}}})\}_{n_{\text{r}}=1}^{{N_{\text{r}}}}$. Here, $\text{Act}^{\text{w/r}}_{n_{\text{w/r}}}$ denote actions associated with specific furniture $\text{Fur}_{n_{\text{w}}}$ or areas $\text{Area}_{n_{\text{r}}}$, while $N_{\text{w}}$ and $N_{\text{r}}$ represent the respective counts of extracted subtasks.
Building on this decomposition, we introduce two specialized progress estimation experts to monitor execution status. 
These modules utilize the hierarchical observations and decomposed tasks to predict the current waypoint execution progress $\mathbf{g}^{\text{w}}_t = \text{Expert}_{\text{prog}}^{\text{w}}(\mathbf{T}_{\text{w}}, \mathbf{h}_t^{\text{w}}, \mathbf{p}_t^{\text{w}})$ and region traversal progress $\mathbf{g}^{\text{r}}_t = \text{Expert}_{\text{prog}}^{\text{r}}(\mathbf{T}_{\text{r}}, \mathbf{h}_t^{\text{r}}, \mathbf{p}_t^{\text{r}})$, respectively.
We employ two specialized experts to reason over these distinct scales. 
First, the waypoint-level reasoning expert $\text{Expert}_{\text{act}}^{\text{w}}$ focuses on local navigability, utilizing object and furniture semantics to predict next action $\mathbf{a}^{\text{w}}_t$,
\begin{equation}
    \mathbf{a}^{\text{w}}_t = \text{Expert}_{\text{act}}^{\text{w}}(\mathbf{T}_{\text{w}},\mathbf{S}_t,\mathbf{R}^{\text{w}}_t, \mathbf{h}^{\text{w}}_t,\mathbf{g}_t^{\text{w}}, \mathbf{D}_t).
\end{equation}
Complementarily, the region expert $\text{Expert}_{\text{act}}^{\text{r}}$ handles global spatial reasoning. 
It aligns the agent's trajectory with panoramic cues to determine the next target region $\mathbf{a}^{\text{r}}_t$, ensuring adherence to structural transitions,
\begin{equation}
    \mathbf{a}^{\text{r}}_t = \text{Expert}_{\text{act}}^{\text{r}}(\mathbf{T}_{\text{r}},\mathbf{S}_t^{\text{r}},\mathbf{R}^{\text{r}}_t, \mathbf{h}^{\text{r}}_t,\mathbf{g}^{\text{r}}_t, \mathbf{D}_t).
\end{equation}
With two experts, our approach grounds navigation in both object-level details and region-structural constraints. However, when the two decisions diverge due to ambiguous observations, relying solely on current views is insufficient. 

\textit{Autonomous Exploration–Based Decision.}
To resolve potential conflicts, we introduce an autonomous exploration strategy governed by a spatial consistency check. 
We define the expert predictions as consistent if the predicted waypoint $\mathbf{a}^{\text{w}}_t$ spatially resides within the predicted region $\mathbf{a}^{\text{r}}_t$. 
When a discrepancy arises, the module identifies a target exploration region $\mathbf{r}_t^{\text{expl}} \in \mathcal{R}$ to gather missing cues:
\begin{equation}
    \mathbf{r}_t^{\text{expl}} = \text{Expert}_{\text{expl}}(\mathbf{I}, \mathbf{S}_t, \mathbf{R}^{\text{r}}_t, \mathbf{g}^{\text{w}}_t, \mathbf{g}^{\text{r}}_t, \mathbf{a}_t^{\text{w}}, \mathbf{a}_t^{\text{r}}).
\end{equation}
Subsequently, we select a navigation goal waypoint $\mathbf{w}^{\text{expl}}_t$ within $\mathbf{r}_t^{\text{expl}}$, defaulting to a LiDAR-based frontier point if the region contains no candidate waypoints.
Upon reaching $\mathbf{w}^{\text{expl}}_t$, the agent acquires supplementary observations $\mathbf{S}^{\text{expl}}_t$ and $\mathbf{R}^{\text{expl}}_t$ to yield the augmented perception sets $\hat{\mathbf{S}}_t = \mathbf{S}_t \cup \mathbf{S}^{\text{expl}}_t$ and $\hat{\mathbf{R}}^{\text{r}}_t = \mathbf{R}_t \cup \mathbf{R}^{\text{expl}}_t$.
The navigation target waypoint $\mathbf{a}_t$ is determined by a conditional policy that utilizes this augmented context to resolve the initial discrepancy, i.e.,
\begin{equation}
    \mathbf{a}_t = 
    \begin{cases} 
    \mathbf{a}^{\text{w}}_t,  \text{if } \text{Consistent}(\mathbf{a}^{\text{w}}_t, \mathbf{a}^{\text{r}}_t); \\
    \text{Expert}_{\text{act}}^{\text{w}}(\mathbf{T}_{\text{w}},\hat{\mathbf{S}}_t,\hat{\mathbf{R}}^{\text{w}}_t, \mathbf{h}^{\text{w}}_t,\mathbf{g}_t^{\text{w}}, \mathbf{D}_t),  \text{otherwise}.
    \end{cases}
\end{equation}
Consequently, we derive the final decision action, which corresponds to the position of the selected candidate waypoint.

\subsection{Real-World Transfer Strategy}
In this section, we discuss the effective sim2real transfer of our method. To ensure a successful transition, we primarily focus on optimizing candidate waypoint selection and enhancing obstacle avoidance capabilities during navigation.

\noindent\textbf{Value-Based Waypoint Sampling}
To mitigate sim-to-real domain gaps where vision-based predictions become unreliable, we propose a heuristic waypoint generation strategy.
We discretize the local environment into a polar grid $\mathbf{M}_t \in \mathbb{R}^{120 \times 12}$, covering a $360^\circ$ field of view and a 3\,m range with angular and radial resolutions of $3^\circ$ and 0.25\,m, respectively.
A score $m_{u,v}$ is assigned to each grid cell $(u, v)$ by integrating geometric navigability with semantic relevance, i.e.,
\begin{equation}
    m_{u,v} = r^{\text{nav}}_{u,v} \cdot \omega^{\text{sem}}_{u,v},
\end{equation}
where $r^{\text{nav}}_{u,v} \in \{0, 1\}$ represents the occupancy state (obstacle or free space) derived from the LiDAR map. To prioritize areas with higher information density, $\omega^{\text{sem}}_{u, v}$ denotes a semantic weight assigned based on the semantic richness at coordinates $(u, v)$. 
To compute $\omega^{\text{sem}}_{u, v}$, we evaluate the confidence scores of the semantic predictions generated by RAM across the twelve discrete views. 
For each view, we enumerate the semantic labels $\mathbf{l}$ that exceed a specific confidence threshold $\delta_{\text{s}}$. 
If the count of high-confidence semantics in the $i-$th view surpasses a hyperparameter $\text{Count}(\mathbf{l}_i)>\delta_{\text{n}}$, the grid cells in $\mathbf{M}_t$ corresponding to that view's angular sector are assigned a value of 2; otherwise, they are set to 0.
Based on $\mathbf{M}$, we sample candidate waypoints by prioritizing distant, high-score regions and merging spatially adjacent candidates. 
Finally, to ensure spatial diversity, we supplement these with random points sampled from free space in angular sectors distant from the selected candidates.

\noindent\textbf{Navigation and Obstacle-Avoidance Strategy}
To achieve real-time, obstacle-aware navigation toward selected position, we employ an end-to-end Deep Reinforcement Learning (DRL) policy $\pi_{\theta}(\cdot)$~\cite{lian2024tdanet} as the low-level motion controller. 
Standard 2D LiDARs are constrained by their inability to detect obstacles outside a fixed height. 
To mitigate this limitation, we enhance the perception module by constructing a comprehensive state representation $\mathbf{G}_t$ via multi-modal sensor fusion. 
To capture obstacles within the vertical field of view, we augment the standard LiDAR map $\mathbf{G}_t^{\text{o}}$ with a depth-derived map $\mathbf{G}_t^{\text{d}}$. 
By integrating these with the agent's position $\mathbf{G}_t^{\text{p}}$ and goal maps $\mathbf{G}_t^{\text{g}}$, we construct the composite state $\mathbf{G}_t=[\mathbf{G^{\text{o}}_t, \mathbf{G}^{\text{d}}_t}, \mathbf{G}^{\text{p}}_t, \mathbf{G}^{\text{g}}_t] \in \mathbb{R}^{3\times H \times W}$. 
Consequently, the policy generates continuous velocity commands $\mathbf{a}^{\text{n}}_t$ as, i.e.,
\begin{equation}
    \mathbf{a}^{\text{n}}_t=\pi_{\theta}(\mathbf{G}_t).
\end{equation}
Adhering to the training paradigm of TDANet~\cite{lian2024tdanet}, we retrain the policy utilizing these augmented inputs. 
Following extensive evaluation, the model demonstrating the optimal performance is selected and deployed on the physical robot to effectively navigate complex real-world obstacles.

\section{Experiments}\label{sec:exp}
To comprehensively evaluate the effectiveness of our proposed method, we conduct extensive experiments in both simulated and real-world environments. 
This section begins by detailing the experimental setup, including datasets, baselines, and evaluation metrics. 
Subsequently, we demonstrate the superiority of our approach through quantitative comparisons against SoTA methods in both simulation and physical scenarios. 
Finally, we present ablation studies to analyze the specific contributions of each core module.
\subsection{Experimental Setup}
\begin{figure*}[!t]
\centering
\includegraphics[width=1\linewidth]{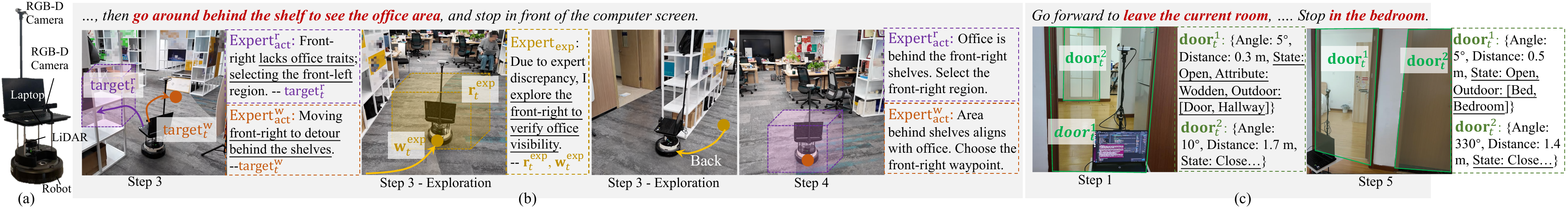}
\caption{Visualization of Spatial-VLN deployed in real-world environments. (a) Configuration of the physical robotic platform. (b) An office navigation sequence involving ambiguous instructions, illustrating the activation of the active exploration mechanism to rectify decision ambiguity. (c) A multi-region navigation task in a home environment involving door interactions, demonstrating how door attribute recognition and regional perception enhancement facilitate complex transitions between distinct areas.}
\label{fig:real}
\end{figure*}



\begin{table}[t]
  \centering
  \caption{Performance comparison with State-of-the-Art supervised and Zero-Shot methods in unseen environments. Note that lower values indicate better performance for NE, whereas higher values denote superior performance for other metrics}
  \resizebox{\linewidth}{!}{
    \begin{tabular}{l|ccccc}
    \toprule
    \multicolumn{1}{c|}{Method}  & NE   & nDTW    & OSR    & SR     & SPL  \bigstrut\\
    \midrule
    \multicolumn{6}{c}{Supervised Methods} \bigstrut\\
    \midrule
    RecBERT~\cite{hong2022bridging}  & 5.80  & 54.81  & 57    & 48    & 43.22 \bigstrut[t] \\
    ETPNav~\cite{an2024etpnav}  & 5.15  & 61.15  & 58    & 52    & 53.18  \bigstrut[b] \\
    \midrule
    \multicolumn{6}{c}{Zero-Shot Methods} \bigstrut\\
    \midrule
    Random  & 8.63  & 34.08  & 12    & 2     & 1.50  \bigstrut[t]\\
    DisscussNav-GPT4~\cite{long2024discuss}   & 7.77  & 42.87  & 15    & 11    & 10.51  \\
    SmartWay-GPT4o~\cite{shi2025smartway} &7.01 &- & 51 & 29 & 22.46 \\
    Open-Nav-Llama3.1~\cite{qiao2025open}   & 7.25  & 44.99  & 23    & 16    & 12.90  \\
    InstructNav-GPT4~\cite{long2024instructnav}  & 9.20 & - & 47 & 17 & 11\\
    CA-Nav-GPT4~\cite{chen2025constraint} & 7.58 & - & 48 & 25.3 & 10.8 \\
    Open-Nav-GPT4~\cite{qiao2025open}   & 6.70  & 45.79  & 23    & 19    & 16.10  \\
    \rowcolor{rowcolor}
    GPT-OSS-20B (Ours)   & 7.33  & 46.95  & 32    & 25    & 21.50 \\
    \rowcolor{rowcolor}
    Deepseekv3 (Ours)   & 6.65  & 47.37  & 34    & 33    & 27.44  \bigstrut[b]\\
    \bottomrule
    \end{tabular}%
    }
  \label{tab:sota}%
\end{table}%

\begin{table}[htbp]
  \centering
  \caption{Comparative analysis of our method and Open-Nav using different LLM experts in unseen environments. `(API)' denotes commercial API access, while others refer to locally deployed models}
  \label{tab:addlabel}
  \resizebox{\linewidth}{!}{
    \begin{tabular}{l|ccccc}
    \toprule
    \multicolumn{1}{c|}{LLM Expert} & {NE} & {nDTW}  & {OSR}  & {SR}  & {SPL}  \\
    \midrule
    \multicolumn{6}{c}{{Open-Nav}} \\
    \midrule
    \hspace{3mm}Llama3.1-70B & 7.25 & 44.99 & 23 & 16 & 12.90 \\
    \hspace{3mm}Qwen2-72B    & 8.14 & 43.14 & 23 & 14 & 12.11 \\
    \hspace{3mm}Gemma-27B    & 6.76 & 40.57 & 16 & 12 & 10.65 \\
    \hspace{3mm}Phi3-14B     & 8.53 & 33.64 & 8  & 5  & 3.81 \\
    \cmidrule(lr){1-6} 
    \hspace{3mm}GPT4~(API)        & 6.70 & 45.79 & 23 & 19 & 16.10 \\
    \hspace{3mm}Deepseek-v3~(API) & 7.17 & 44.95 & 23 & 16 & 13.11 \\
    \midrule
    \multicolumn{6}{c}{{Spatial-VLN (Ours)}} \\
    \midrule
    \hspace{3mm}Llama3.1-70B & 7.54 & 41.40 & 20 & 19 & 15.36 \\
    \hspace{3mm}Qwen2-72B    & 7.64 & 44.85 & 22 & 18 & 15.53 \\
    \hspace{3mm}Gemma-27B    & 7.80 & 43.80 & 25 & 23 & 18.90 \\
    \hspace{3mm}GPT-Oss-20B  & 7.33 & 46.95 & 32 & 25 & 21.50 \\
    \hspace{3mm}Phi3-14B     & 8.68 & 35.37 & 9  & 7  & 6.18 \\
    \cmidrule(lr){1-6}
    \hspace{3mm}Deepseek-v3~(API)    & 6.65 & 47.37 & 34 & 33 & 27.44 \\
    \hspace{3mm}Doubao-1.5~(API) & 6.78 & 50.76 & 34 & 26 & 23.11 \\
    \bottomrule
    \end{tabular}%
  }
  \label{tab:llm}%
\end{table}

\noindent\textbf{Implementation Details.} 
Simulation experiments are conducted on the Habitat platform, utilizing LLM experts via either external APIs or local deployment. 
The hardware setup includes a single NVIDIA RTX 4090 GPU for standard evaluation, scaled to a dual-GPU configuration for local LLM inference. 
For real-world experiments, as shown in~\Cref{fig:real}(a), we employ a Turtlebot 4 mobile robot. 
The perception suite consists of a 360$^\circ$ LiDAR, a RealSense camera mounted at a height of 1.3\,m for panoramic RGB-D observations, and an OAK camera positioned at 0.4\,m to generate depth maps for obstacle avoidance. 
All onboard computing is handled by a laptop equipped with a single RTX 4090 GPU,  mounted on the robot. 
RAM, Spatialbot, and proposed Spatial-VLN run locally on the laptop, while accessing LLM experts through online APIs or local server requests.
For the semantic weight calculation during waypoint generation, we utilize a semantic confidence threshold of $\delta_{\text{s}} = 0.85$ and a minimum high-confidence count threshold of $\delta_{\text{n}}=1$.
 
\noindent\textbf{Datasets.} The simulation benchmarks, the challenge-specific subsets, and the real-world evaluation scenarios used in the experiments are detailed below.

\textit{Simulation Benchmark.} We conduct our simulation experiments on VLN-CE benchmark, which utilizes the Habitat simulator~\cite{2019habitat} to render continuous trajectories and is built upon the Room-to-Room (R2R) dataset~\cite{anderson2018vision} with 90 distinct scenes from Matterport3D (MP3D)~\cite{anderson2018vision}. To compare generalization performance against supervised training methods, we follow existing zero-shot methods evaluating our method on the unseen environment dataset.

\textit{Challenge Dataset Selection.} To evaluate the model's robustness against the three spatial challenges defined earlier, we implement a challenge difficulty scoring mechanism to select the top-100 episodes with the highest cumulative complexity. Concretely, for Doorway Interaction and Multi-Region tasks, we quantify complexity by calculating the frequency of door-related terms (e.g., `door') and region-related key word (e.g., bedroom', `hallway'), respectively; a higher frequency indicates a greater number of required spatial transitions. Simultaneously, we assess the challenge of Ambiguous Instructions based on the density of specific landmark semantics. Here, a lower proportion of landmark terms implies fewer explicit visual anchors, thus intensifying the difficulty of resolving instructional ambiguity.

\textit{Real-World Scenarios.} To validate Sim-to-Real transfer, we deploy the agent in public offices and homes with diverse room layouts and rich semantic objects. A balanced test set of 40 natural language instructions is used, with predefined start and goal configurations. Dynamics obstacles, including unrestricted pedestrian movements, are introduced to emulate realistic deployment conditions. 

\noindent\textbf{Evaluation Metrics.}
To quantitatively assess navigation performance, we employ standard metrics for the simulation benchmarks, including Trajectory Length (TL), Navigation Error (NE), Oracle Success Rate (OSR), Success Rate (SR), and Success weighted by Path Length (SPL). However, for real-world evaluation, we impose a success criterion stricter than the standard simulation metric which solely relies on a geometric threshold of stopping within 3 meters. Crucially, we incorporate an additional spatial-semantic verification to prevent false positives where the agent is geometrically close but semantically incorrect. For instance, in a task instructed to `stop in the kitchen', a trial is deemed a failure if the robot stops within the 3-meter radius but remains in the adjacent corridor rather than entering the target semantic region.

\subsection{Simulation Environmental Results}
\label{sec:sim}
Simulation results are presented to evaluate Spatial-VLN against SoTA baselines, and assess its  generalization across LLM models and its robustness in challenging spatial tasks.

\textit{Comparative Analysis with State-of-the-Arts.} We first benchmark Spatial-VLN against SoTA methods in the VLN-CE domain. Our comparison includes supervised approaches, such as RecBERT~\cite{hong2022bridging} and ETPNav~\cite{an2024etpnav}, which leverage end-to-end training and topological graphs to establish strong baselines using learned navigation priors. We also compare against Zero-Shot methods harnessing Large Language Models, including NavGPT~\cite{zhou2024navgpt}, and Open-Nav~\cite{qiao2025open}, which serves as the primary baseline due to its adaptability across LLM model scales. As shown in~\Cref{tab:sota}, Spatial-VLN achieves a superior 33\% success rate. Notably, our framework outperforms methods relying on GPT-4 even when utilizing lower-cost APIs or locally deployed models, indicating that specialized spatial reasoning is more effective for navigation than simply increasing raw inference power.

\textit{Generalization Across LLM.} Table~\Cref{tab:llm} details Spatial-VLN performance across different LLMs. The DeepSeek-v3 API achieves the highest 33\% success rate, surpassing the Open-Nav baseline by 17\% under identical configurations. Medium-sized local models also exhibit robust performance, where Gemma-27B exceeded Open-Nav by 11\% in success rate. These results validate the framework's strong generalization capabilities and its operational feasibility in both high-end and resource-constrained environments.

\textit{Performance on Challenging Spatial Tasks.} To investigate model behavior under demanding conditions, we evaluate performance across three challenge tasks in~\Cref{tab:sim-chan}. Spatial-VLN consistently leads the baseline, achieving gains of 12\% in doorway interaction and 11\% in ambiguous instruction tasks. These improvements result from fine-grained perception of door attributes and the active exploration mechanism. The more marginal gains in multi-region tasks likely reflect the inherent difficulty of long-horizon navigation regarding long-term memory and progress estimation.



\label{sec:set}
\begin{figure}[!t]
\centering
\includegraphics[width=1\linewidth]{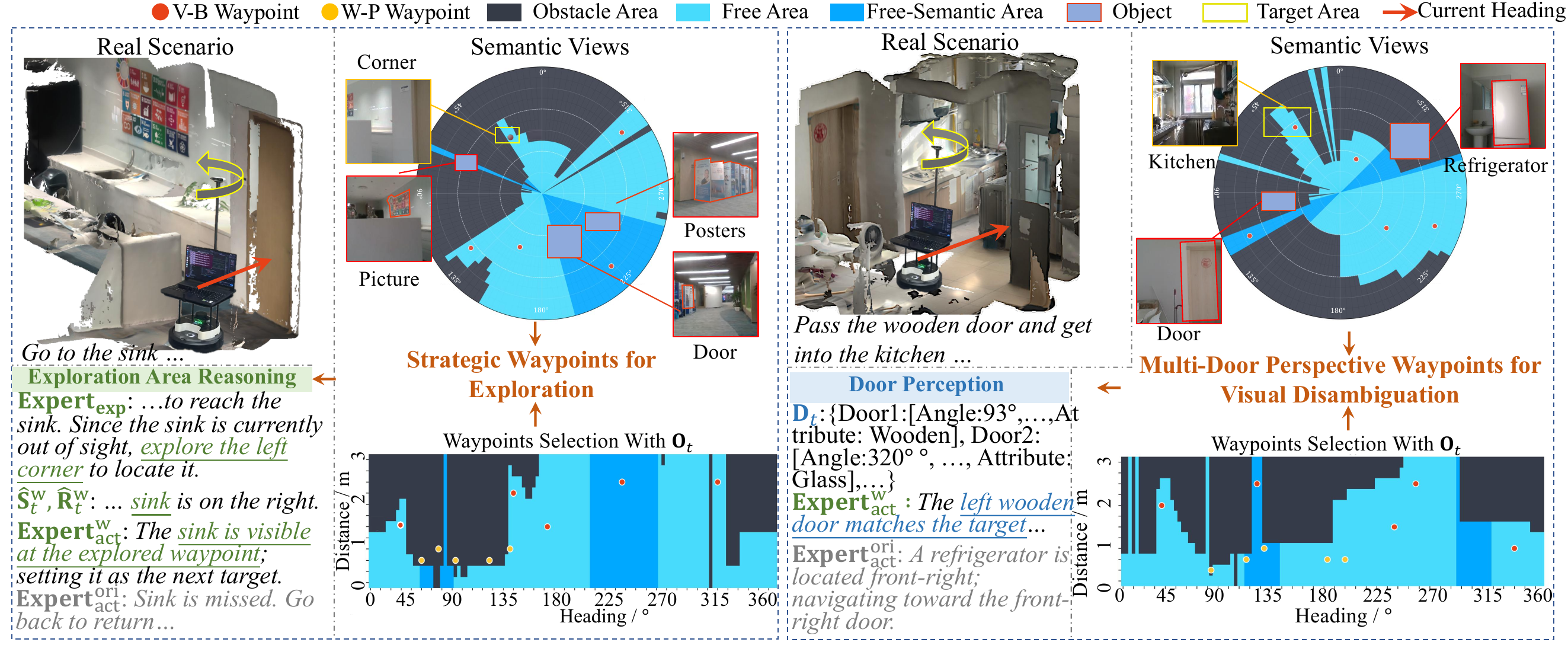}
\caption{{Visualization of candidate waypoint distribution in real-world environments. Value-based maps, incorporating free-space availability and semantic richness outperform traditional candidate predictors in real-world scenarios. These value-sampled waypoints provide reliable navigation targets and viewpoints for the dual modules of Spatial-VLN, facilitating effective sim-to-real transfer.} }
\label{v-waypoint}
\end{figure}

\begin{table}[!t]
  \centering
  \caption{Performance comparison between Open-Nav and Spatial-VLN across three challenging tasks in simulated unseen environments. `DI', `MR', and `AI' denote Doorway Interaction, Multi-Room, and Ambiguous Instruction tasks, respectively}
    \begin{tabular}{c|c|cccccc}
    \toprule
    Method & Task  & TL    & NE    & nDTW  & OSR   & SR    & SPL \bigstrut\\
    \midrule
    \multirow{3}[2]{*}{Open-Nav} & DI & 6.60  & 8.05  & 40.99  & 12    & 10    & 7.67  \bigstrut[t]\\
          & MR & 6.58  & 8.28  & 40.74  & 17    & 14    & 11.41  \\
          & AI & 6.05  & 6.82  & 48.36  & 26    & 12    & 10.60  \bigstrut[b]\\
    \hline
    \multirow{3}[2]{*}{\shortstack{Ours}} & DI & 6.08  & 7.08  & 48.47  & 25    & 22    & 19.95  \bigstrut[t]\\
          & MR & 6.14  & 8.24  & 42.27  & 17    & 16    & 14.02  \\
          & AI & 6.02  & 6.76  & 47.26  & 28    & 25    & 21.93  \bigstrut[b]\\
    \bottomrule
    \end{tabular}%
  \label{tab:sim-chan}%
\end{table}%

\subsection{Real-World Experimental Results}
\label{sec:real}
To validate the proposed Spatial-VLN framework, we conduct extensive experiments in both office and home environments and provide a comprehensive performance analysis.

\textit{Quantitative Performance in Real-World Challenges.} We evaluate Spatial-VLN against the Open-Nav baseline using 40 instructions under identical settings. As shown in \Cref{tab:real}, Spatial-VLN outperforms the baseline across all metrics. Specifically, our method achieves an average SR (Avg-SR) of 44\% and 52\% in Office and Home scenes respectively, substantially surpassing Open-Nav's 28\% and 34\%, while also reducing navigation errors. 
Notably, the performance in real-world experiments slightly exceeds that in simulation. This is primarily attributable to the visual fidelity of physical environments; while simulation suffers from reconstruction artifacts like broken meshes and blurred textures, real-world scenes offer clear, continuous visual cues. Combined with our enhanced perception module, this enables more accurate identification of structural features such as doorframes, leading to robust navigation.

\textit{Qualitative Visualization of Navigation Behaviors.} To further validate the robustness of our method, we provide qualitative visualizations of the navigation process. As illustrated in~\Cref{fig:real}(b)-(c), when facing instructions with spatial ambiguity, the robot actively explores the uncertain regions to rectify the ambiguity. The visualization also highlights the efficiency of our SPE module; for instance, when encountering multiple doors, the agent distinguishes specific door attributes to select the correct exit for cross-region tasks. These capabilities significantly strengthen the alignment between current observations and tasks and improve navigation performance.

\textit{Visualization of Candidate Waypoint Distribution.} {Additionally, we analyze the candidate waypoint sampling strategy in \Cref{v-waypoint}.} Unlike the baseline predictor that tends to generate clustered, short-range candidates or place them within obstacles, our strategy samples reachable, long-distance waypoints within semantically rich free space. This distribution effectively mitigates common issues such as non-navigability, overlapping fields of view, and semantic sparsity, ensuring safer navigation steps with more semantic information.


\begin{table}[t]
  \centering
  \caption{Performance comparison between Open-Nav and Spatial-VLN across three challenge tasks in real-world environments. A-SR and A-NE denote the average SR and NE across different scenes, respectively}
  \label{tab:real}
  \setlength{\tabcolsep}{5pt} 
  \resizebox{\linewidth}{!}{
  \begin{tabular}{lcccccc}
    \toprule
    Method & Scene & DI (SR) & MR (SR) & AI (SR) & A-SR & A-NE $\downarrow$ \\
    \midrule
    Open-Nav & Office & 0.29 & 0.14 & 0.40 & 0.28 & 3.24 \\
    {Ours} & Office & {0.43} & {0.29} & {0.60} & {0.44} & {2.93} \\
    \midrule
    Open-Nav & Home & 0.40 & 0.38 & 0.25 & 0.34 & 2.80 \\
    {Ours} & Home & {0.60} & {0.50} & {0.46} & {0.52} & {2.11} \\
    \bottomrule
  \end{tabular}}
\end{table}

\begin{table}[t]
  \centering
  \caption{Ablation study on the contribution of Spatial Perception Enhancement module (SPE) and Explored Multi-expert Reasoning module (EMR) across three challenge tasks in simulated unseen environments}
    \begin{tabular}{c|c|c|cccccc}
    \toprule
    \multirow{2}[4]{*}{Task} & \multicolumn{2}{c|}{Method} & \multirow{2}[4]{*}{TL} & \multirow{2}[4]{*}{NE} & \multirow{2}[4]{*}{nDTW} & \multirow{2}[4]{*}{OSR} & \multirow{2}[4]{*}{SR} & \multirow{2}[4]{*}{SPL} \bigstrut\\
\cline{2-3}          & SPE & EMR &       &       &       &       &       &  \bigstrut\\
    \midrule
    \multicolumn{1}{c|}{\multirow{2}[2]{*}{DI}} & \ding{51}     & \ding{55}     & 6.10  & 7.06  & 47.56  & 20    & 16    & 14.16  \bigstrut[t]\\
          & \ding{55}     & \ding{51}     & 6.09      & 7.16      & 47.34      & 23      & 19      & 16.98 \\
    \hline
    \multicolumn{1}{c|}{\multirow{2}[2]{*}{MR}} & \ding{51}     & \ding{55}     & 6.14  & 8.01  & 45.00  & 18    & 15    & 12.77  \bigstrut[t]\\
          & \ding{55}     & \ding{51}     & 6.14      & 8.15      & 43.37      & 15      & 15      & 13.07 \\
    \hline
    \multicolumn{1}{c|}{\multirow{2}[2]{*}{AI}} & \ding{51}     & \ding{55}     & 6.02  & 6.99  & 46.60  & 23    & 19    & 16.51  \bigstrut[t]\\
          & \ding{55}     & \ding{51}     & 7.64      &  6.66     & 47.00      & 25      &  23     & 19.55 \\
    \bottomrule
    \end{tabular}%
  \label{tab:abl}%
\end{table}%
\subsection{Ablation Studies}
\label{sec:ablation}
Ablation studies in~\Cref{tab:abl} verify the contributions of the SPE and EMR modules, with single-module evaluations revealing distinct functional dependencies. In ambiguous instruction and door interaction tasks, the EMR module consistently outperforms SPE with SR gains of 3\% and 4\% respectively, indicating that exploration provides vital supplementary information. However, these variants still trail the full Spatial-VLN, proving that exploration requires perceptual grounding to achieve optimal performance. In multi-region tasks, both single-module variants stagnate at 15\% SR and underperform the complete framework. These findings confirm that neither perception nor reasoning alone suffices for complex navigation, necessitating their synergistic integration.

\vspace{-1mm}
\section{Conclusion}
To address the deficiency of spatial reasoning in LLM-based zero-shot VLN-CE, this paper identified three spatial challenges and proposed the Spatial-VLN framework. This framework integrated a spatial perception enhancement module to ensure structural consistency and an explored multi-expert reasoning module that triggered active exploration upon decision conflicts. Leveraging value-based waypoint sampling and a DRL-based policy for sim-to-real transfer, the proposed approach achieved state-of-the-art performance in both simulated and real-world environments, as demonstrated through extensive comparison experiments. Future work will focus on robust long-term memory architectures to further enhance navigation performance in complex and long-horizon scenarios.

\bibliographystyle{IEEEtran} 
\bibliography{reference}

\begin{IEEEbiography}
[{\includegraphics[width=1in,height=1.25in,clip,keepaspectratio]{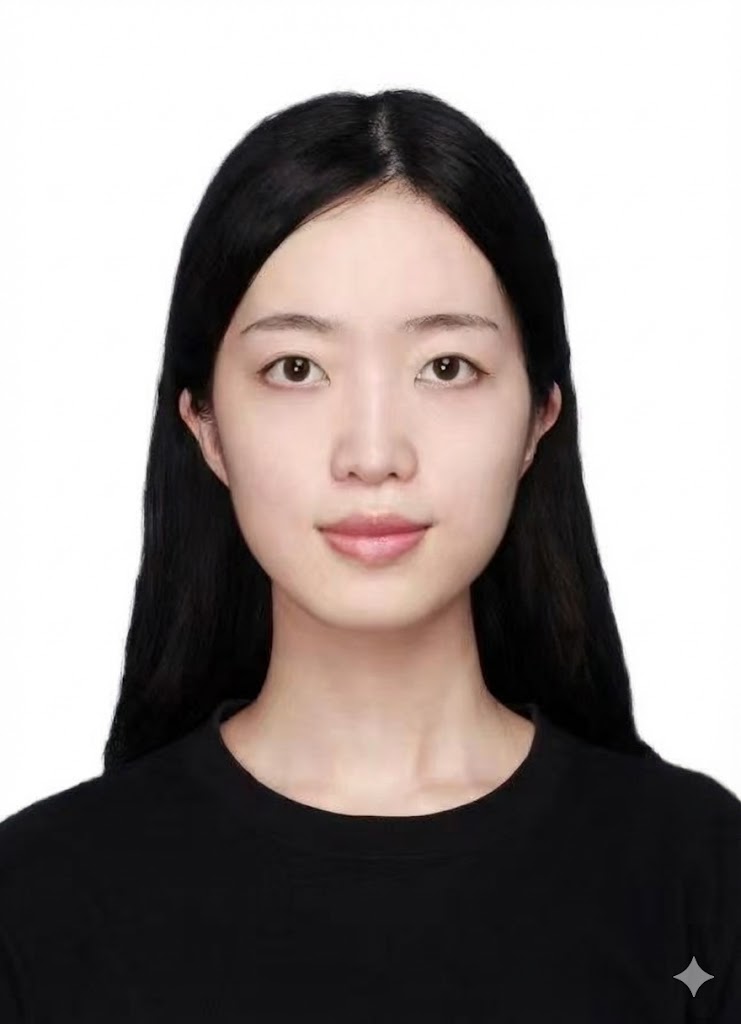}}]{Lu Yue} 
received the B.S. and M.S. degrees from the School of Control Science and Engineering, Shandong University, Jinan, China, in 2019 and 2022, respectively. She is currently working toward the Ph.D. degree with the School of Advanced Manufacturing and Robotics, Peking University, Beijing, China. 

Her research interests include embodied navigation, deep learning, and robot learning.
\end{IEEEbiography}

\begin{IEEEbiography}
[{\includegraphics[width=1in,height=1.25in,clip,keepaspectratio]{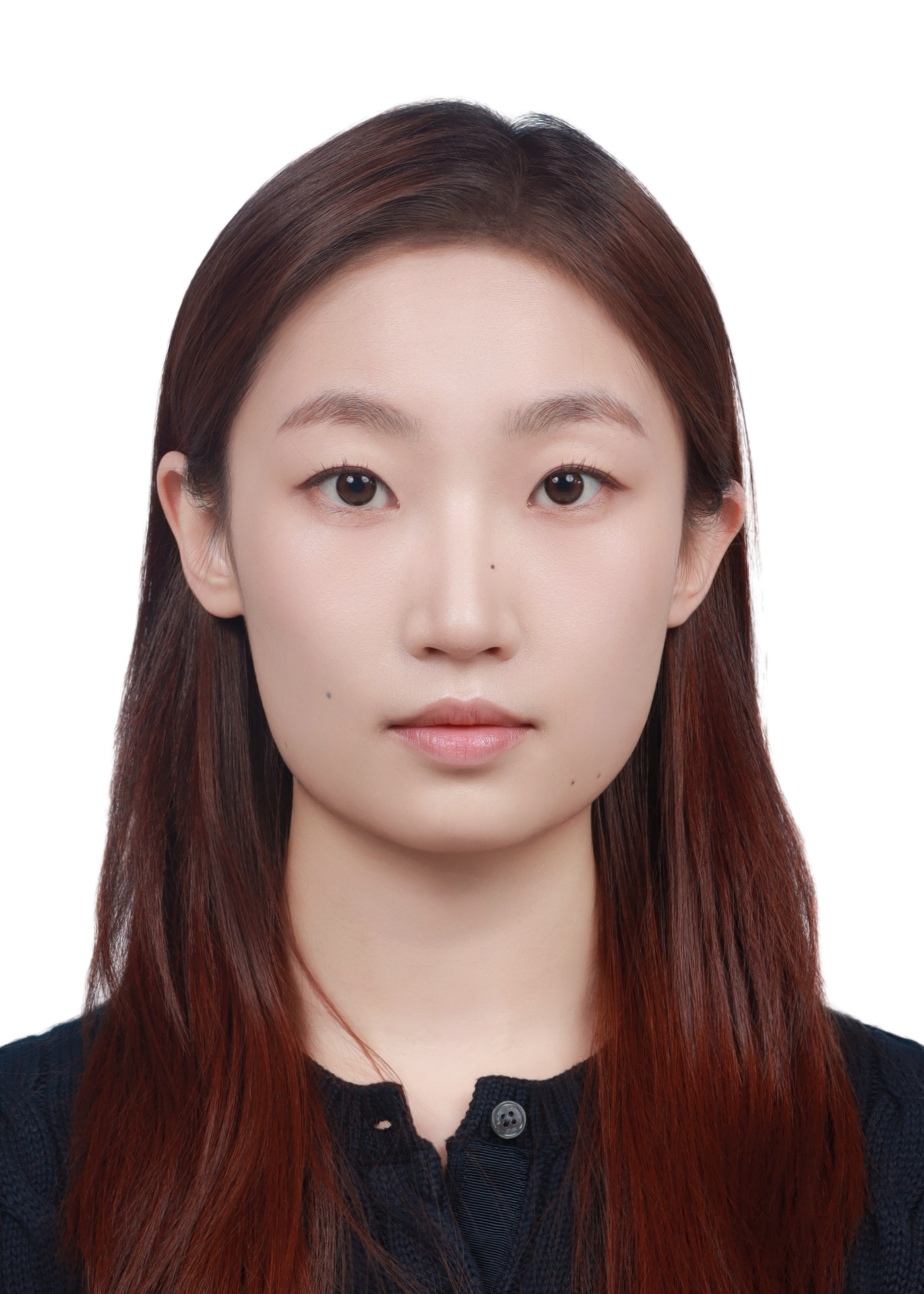}}]{Yue Fan} 
received the B.S. degree in Software Engineering from China University of Geosciences (Beijing), Beijing, China, in 2023. She is currently working toward the M.S. degree with the School of Advanced Manufacturing and Robotics, Peking University, Beijing, China. 

Her research interests include embodied navigation, deep learning, and robot learning.
\end{IEEEbiography}

\begin{IEEEbiography}
[{\includegraphics[width=1in,height=1.25in,clip,keepaspectratio]{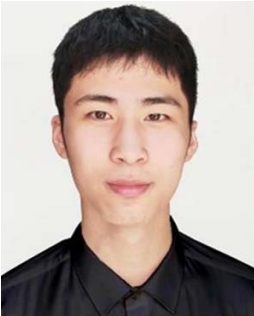}}]{Shiwei Lian} 
received the B.S. degree in mechatronic engineering from Tongji University,Shanghai, China, in 2022, and the M.S. degree in mechanical engineering with the College of Engineering, Peking University, Beijing, China.

His research interests include deep reinforcement learning and autonomous navigation.
\end{IEEEbiography}

\begin{IEEEbiography}
[{\includegraphics[width=1in,height=1.25in,clip,keepaspectratio]{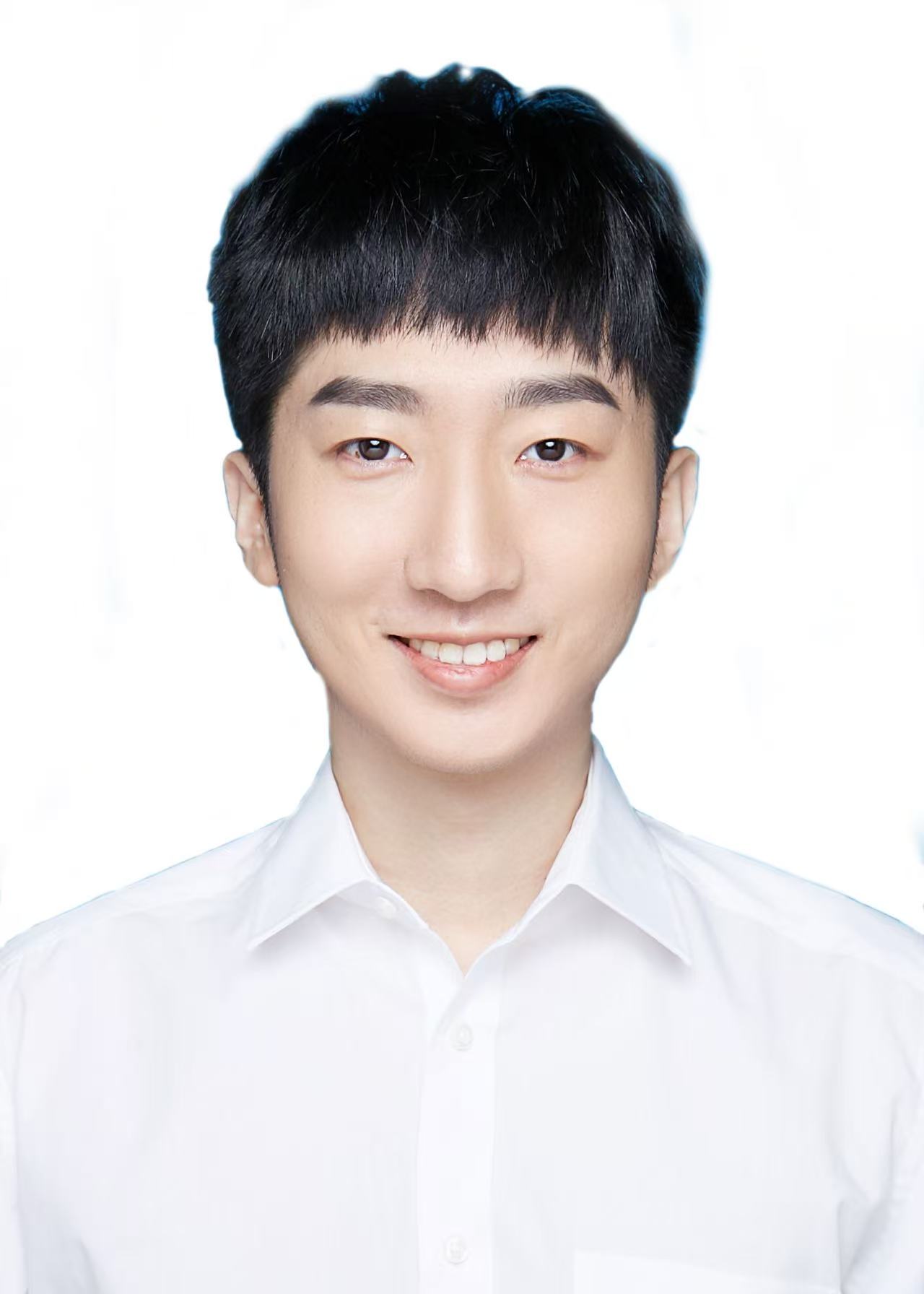}}]{Yu Zhao} 
received the B.S. degree from Soochow University, Suzhou, China, in 2012, the M.S. degree from Southeast University, Nanjing, China, in 2015, and the Ph.D. degree from Tianjin University, Tianjin, China, in 2024. He was a Visiting Researcher with the NExT++ Research Centre, National University of Singapore, from 2023 to 2024. He is currently a Postdoctoral Research Associate with the Tianjin Artificial Intelligence Innovation Center and the Harbin Institute of Technology (Shenzhen), Shenzhen, China. 

His research interests include natural language processing and intelligent human-computer interaction.
\end{IEEEbiography}

\begin{IEEEbiography}
[{\includegraphics[width=1in,height=1.25in,clip,keepaspectratio]{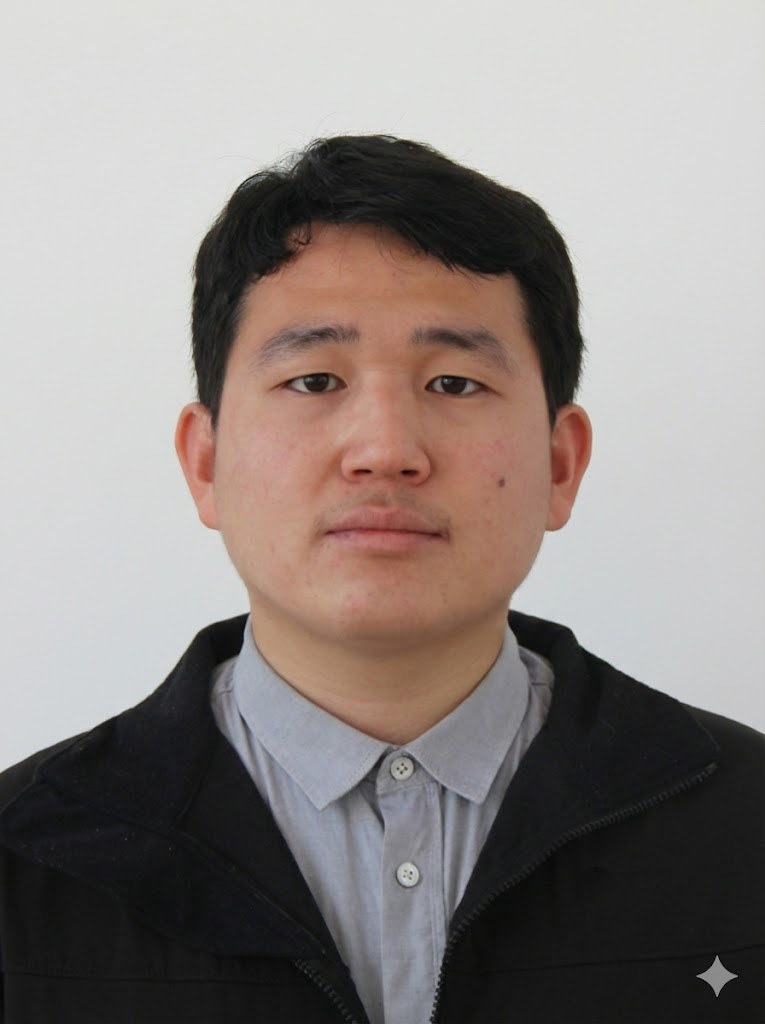}}]{Jiaxin Yu} 
 is currently an Assistant Researcher at the Defense Innovation Institute, Academy of Military Sciences (AMS). 
 
 His research interests include human–computer interaction (HCI) and 3D spatial interaction, with a particular focus on multimodal interaction and user interface evaluation in virtual reality.
\end{IEEEbiography}

\begin{IEEEbiography}
[{\includegraphics[width=1in,height=1.25in,clip,keepaspectratio]{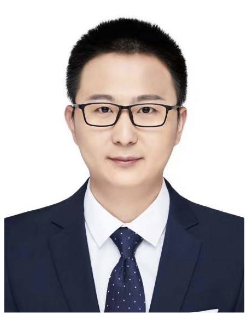}}]{Liang Xie} 
received the B.S., M.S., and Ph.D. degrees from the National University of Defense Technology, Changsha, China, in 2012, 2014, and 2018, respectively. He is currently an Assistant Researcher with the National Innovation Institute of Defense Technology, Academy of Military Sciences, China, Beijing, China. 

His research interests include computer vision, human–machine interaction, and mixed reality.
\end{IEEEbiography}

\begin{IEEEbiography}
[{\includegraphics[width=1in,height=1.25in,clip,keepaspectratio]{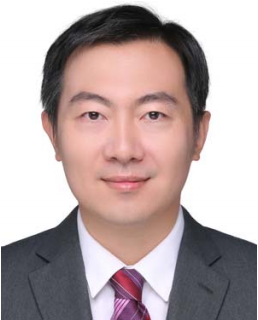}}]{Feitian Zhang}
received the Bachelor's and Master's degrees in Automatic Control from Harbin Institute of Technology, Harbin, China, in 2007 and 2009, respectively, and the Ph.D. degree in Electrical and Computer Engineering from Michigan State University, East Lansing, MI, in 2014. 

He is currently an Associate Professor with the School of Advanced Manufacturing and Robotics, Peking University, Beijing, China. 
Prior to joining Peking University, he was an Assistant Professor in the Department of Electrical and Computer Engineering at George Mason University (GMU), Fairfax, VA, and the founding director of the Bioinspired Robotics and Intelligent Control Laboratory (BRICLab) from 2016 to 2021. 
His research interests include bioinspired robotics, control systems, artificial intelligence, underwater vehicles, and aerial vehicles. 
\end{IEEEbiography}

\vfill

\end{document}